\documentclass[12pt]{article}

\usepackage[utf8]{inputenc}
\usepackage[T1]{fontenc}
\usepackage{amsmath, amssymb}
\usepackage{graphicx}
\usepackage{cite}
\usepackage[final,expansion=false]{microtype}
\usepackage{textcomp}
\usepackage{placeins}
\usepackage{geometry}
\usepackage{float}
\usepackage{hyperref}

\title{Geometry Is Not Robustness: A Trajectory-Level Study of PGD Evaluation}
\author{Dhairysheel Durgule\\
\small Independent Researcher\\
\small \href{mailto:dhairysheelsdurgule@gmail.com}{dhairysheelsdurgule@gmail.com}}
\date{}

\begin{document}
\maketitle

\begin{abstract}
Projected Gradient Descent (PGD) is widely used as a standard adversarial attack for evaluating the robustness of deep learning models. Robustness is typically assessed through final adversarial accuracy, which does not capture the dynamic behaviour of models throughout the attack process. Recent work has proposed trajectory-level diagnostics – such as loss evolution, gradient alignment, and steps-to-failure – to provide deeper insight into adversarial optimisation dynamics. However, the extent to which these diagnostics reliably indicate robustness strength remains unclear.

In this work, we conduct a trajectory-level investigation of PGD attacks on convolutional neural networks trained on the Fashion-MNIST dataset. We compare clean-trained and adversarially-trained models across multiple robustness regimes under rigorous 20-step PGD evaluations with random initialisation and multiple restarts for robustness measurement, and single-initialisation trajectory recording for diagnostic analysis. By systematically recording full PGD trajectories across 3000 clean-correct samples per model, we analyse loss evolution, gradient alignment, and failure timing behaviour across attack iterations.

Our results reveal a clear robustness hierarchy across models; however, trajectory metrics do not contribute equally to its identification. Mean loss trajectories and gradient alignment patterns appear quantitatively similar across adversarially-trained models with substantially different robust accuracies. In contrast, steps-to-failure distributions provide a clearer separation of robustness regimes by directly reflecting functional resistance to adversarial perturbation.

These findings indicate that trajectory-level diagnostics describe aspects of optimisation geometry but do not independently measure adversarial robustness. Their interpretability depends on robustness regime, attack strength, and multi-metric evaluation. Rather than replacing standard robustness measurements, trajectory-level analysis should be regarded as a complementary diagnostic tool whose signals must be interpreted in context.

\end{abstract}

\section{Introduction}
\subsection{The Issues Connected to Machine Learning Models}

Machine learning models, especially deep learning models, have achieved high accuracy and are widely used in applications such as image classification, speech recognition, and automated decision-making. However, as these systems are increasingly deployed in real-world settings, their robustness under adversarial conditions remains a primary concern. Research has shown that models are highly sensitive to small, carefully designed perturbations that are often imperceptible to the human eye \cite{szegedy2013intriguing}. Despite being visually indistinguishable from the original input, such perturbations can cause models to make incorrect predictions with high confidence.

Adversarial examples exploit gradient information to craft perturbations that manipulate the decision boundary of a model \cite{goodfellow2014explaining}. Their transferability across architectures and datasets suggests that adversarial vulnerability is a systematic characteristic of modern deep networks rather than a mere implementation defect \cite{szegedy2013intriguing}. These findings raise important questions regarding how robustness should be evaluated.

\subsection{Understanding the Nature of Projected Gradient Descent (PGD)}

Projected Gradient Descent (PGD) is one of the most widely used methods for generating adversarial examples. It is an iterative gradient-based attack that repeatedly updates the input to increase the model’s loss while constraining the perturbation size \cite{madry2017towards}. Due to its strength and reliability, PGD has become a standard tool for evaluating adversarial robustness. In practice, robustness is typically measured by adversarial accuracy under a fixed PGD attack, alongside clean accuracy.

However, reducing robustness evaluation to a single scalar metric, final adversarial accuracy, hides important information about how models behave during the attack process. A PGD attack produces a sequence of intermediate inputs corresponding to successive optimisation steps. Two models with similar final robustness accuracy may exhibit very different behaviours along this trajectory. One model may resist the attack for many steps before failing, while another may collapse immediately. Although final accuracy would appear identical, the dynamics of failure may differ substantially, indicating qualitatively different robustness behaviour.

Recent analyses have therefore begun to examine trajectory-level properties of adversarial attacks \cite{ilyas2019adversarial}, including loss evolution and gradient alignment across iterations. These diagnostics aim to provide insight into optimisation dynamics beyond a single summary metric. However, the interpretability of trajectory-level metrics remains contingent on the robustness regime being examined. In certain contexts, trajectory smoothness or stable gradient alignment may occur despite significant differences in robustness strength. Without careful conditioning and multi-metric interpretation, trajectory-based analyses may yield incomplete or misleading conclusions.

\subsection{Motivation and Goals of the Study}

This work analyses complete PGD trajectories to examine when trajectory-level metrics provide meaningful insight into adversarial resilience and when they fail to do so. Using recorded PGD trajectories, we investigate how loss values, gradient alignment, and prediction outcomes evolve across attack iterations. The study conditions trajectory analysis on clean-correct samples and compares clean-trained and adversarially-trained models across multiple robustness regimes.

The central question is whether trajectory geometry alone can reliably distinguish robustness strength. While trajectory-level diagnostics describe optimisation behaviour, their interpretability depends on the model regime and experimental context. Rather than proposing a new attack or defence mechanism, this work adopts a diagnostic perspective on adversarial evaluation and provides practical considerations for interpreting trajectory-based robustness analyses.

\section{Literature Review}
\subsection{Adversarial Examples and Adversarial Robustness}

Deep neural networks exhibit exceptional state-of-the-art performance across various tasks. However, their trustworthiness is compromised by the presence of adversarial examples. Adversarial examples are inputs which are modified by small and carefully crafted perturbations that are imperceptible by humans, yet result in significant misclassification with high confidence \cite{szegedy2013intriguing}. The phenomenon was first systematically documented by Szegedy et al.~\cite{szegedy2013intriguing}, who showed that even state-of-the-art image classifiers could be sensitive to small perturbations and that such adversarial examples transfer across different model architectures and training datasets. This suggests that adversarial vulnerability is a systematic and widespread property of modern deep learning systems rather than an isolated implementation issue.

Building on this discovery, Goodfellow et al.~\cite{goodfellow2014explaining} (2014) proposed that the approximately linear behaviour of neural networks in high-dimensional input spaces is a primary factor contributing to their vulnerability to adversarial attacks. This perspective highlights that even minor adjustments aligned with the gradient of the loss function can accumulate to produce significant alterations in the model's output. The significance of adversarial robustness has emerged as a vital area of research due to the understanding that adversarial examples stem from the inherent geometry of models, rather than being merely a byproduct of overfitting or insufficient generalisation.

\subsection{Gradient-Based Attacks and Projected Gradient Descent}

Following the discovery of adversarial examples, research has focused on developing methods for generating perturbations in a more powerful and efficient way. Among these approaches, gradient-based attacks emerged as a dominant approach due to their efficiency and simplicity. Goodfellow et al.~\cite{goodfellow2014explaining} introduced the Fast Gradient Sign Method (FGSM), a single-step attack that perturbs the input in the direction of the sign of the gradient of the loss with respect to the input. Although this attack is computationally inexpensive, it was later demonstrated to underestimate worst-case adversarial vulnerability.

To address these limitations, Madry et al.~\cite{madry2017towards} proposed Projected Gradient Descent (PGD). It is an iterative attack that extends FGSM by applying multiple gradient ascent steps, with each step projecting the perturbed input back into a constrained perturbation set. PGD was formulated within a robust optimisation framework, where robustness is defined as a min-max problem between the model parameters and an adversarial perturbation. The authors argued that PGD functions are a “universal” first-order adversary, as multiple random initialisations consistently converge to similarly high-loss solutions, indicating that PGD reliably approximates the worst-case adversarial perturbation within the first-order threat model~\cite[p.~6]{madry2017towards}.

Further studies have examined the dynamics of gradient-based attacks. Dong et al.~\cite{dong2018momentum} demonstrated that incorporating momentum into iterative attacks could stabilise gradient directions across steps and improve attack effectiveness, highlighting that the attack trajectory itself influences attack performance.

\subsection{Adversarial Robustness Evaluation Practices}

Standardised evaluation procedures have developed as adversarial attacks have evolved to become more complex. Madry et al.~\cite{madry2017towards} popularised the terms adversarial accuracy or robust accuracy, defined as a model's accuracy under a fixed-step PGD attack applied to the test set. Reporting robust accuracy alongside clean accuracy has become a standard benchmark for measuring model robustness and enables comparisons across architectures and training procedures.

However, such robustness evaluations are particularly sensitive to attack configurations and implementation details. Athalye et al.~\cite{athalye2018obfuscated} highlighted gradient masking as a failure mode where defences appear strong by concealing gradients rather than addressing underlying vulnerabilities, resulting in potentially deceptive robustness claims. To mitigate such issues, Croce \& Hein~\cite{croce2020reliable} introduced AutoAttack, a parameter-free ensemble of complementary attacks designed to provide a more reliable method for robustness evaluation. Their findings demonstrated that under more stringent evaluation, the performance of many previously reliable models significantly declined.

\subsection{Trajectory-Level and Dynamic Analyses of Adversarial Attacks}

Currently, most robustness evaluations rely on final adversarial accuracy, which overlooks the fundamental property of iterative attacks such as PGD that generate a sequence of intermediate inputs forming an attack trajectory. These intermediate inputs correspond to successive optimisation steps taken by the attack algorithm. Analysing these trajectories can provide additional insights into the attack process, including how loss values, gradients, and predictions evolve across steps.

Liu et al.~\cite{liu2020loss} analysed the loss landscape associated with adversarial training and stressed that adversarial objectives can exhibit more complicated and irregular geometry than clean objectives, affecting convergence and gradient behaviour. In support of this view, Ilyas et al.~\cite{ilyas2019adversarial} illustrated that iterative attacks systematically leverage non-robust features that, while highly predictive, are also fragile, thereby explaining how adversarial examples emerge progressively. This process provides insight into how adversarial examples develop across successive optimisation steps.

\subsection{Limitations and Open Questions}

Despite progress in robustness evaluation, interpreting trajectory-level diagnostics remains challenging. Most existing work assumes sufficiently high clean accuracy and implicitly treats trajectory behaviour as intrinsically meaningful. However, little work explicitly examines regimes in which clean accuracy is severely degraded or where clean-correct samples become statistically sparse.

In such regimes, trajectory metrics may exhibit smooth or stable behaviour while reflecting degenerate failure dynamics rather than genuine robustness structure. The present study avoids this failure mode by conditioning trajectory analysis strictly on clean-correct samples across models with clean accuracies between 84\% and 92\%, ensuring that reported trajectory behaviours reflect adversarial vulnerability rather than pre-existing classification errors. The broader question of how trajectory diagnostics behave under severely degraded clean accuracy remains an open direction for future work.

Addressing this interpretability gap, rather than proposing a new attack or defence, forms the central focus of the present study.

\section{Methodology}
\subsection{Dataset}
All experiments were conducted using the Fashion-MNIST dataset \cite{Xiao2017FashionMNISTAN}. It is a standardised benchmark designed for image classification tasks. The dataset comprises 70,000 grayscale images with a resolution of $28 \times 28$ pixels across 10 clothing classes. It was presented as a challenging substitute for the original MNIST dataset while preserving similar structure and scale. Fashion-MNIST is widely used in adversarial robustness studies due to its moderate complexity and well-studied baseline performance, making it suitable for controlled diagnostic analysis of adversarial attacks. The dataset was split using the standard author-provided partition: 60,000 training images and 10,000 testing images.

All images were converted to floating-point tensors in the range $[0,1]$ using standard preprocessing. No additional input normalisation (e.g., mean subtraction or standard deviation scaling) was applied. All gradient computations during training and adversarial attacks were therefore taken with respect to raw pixel values in the $[0,1]$ domain.

\subsection{Model Architecture}
This study utilises a compact convolutional neural network (CNN), referred to as CNNSmall, for image classification on the Fashion-MNIST dataset. The architecture consists of two convolutional layers (32 and 64 channels, respectively), each followed by ReLU activation and max-pooling, and a final fully connected layer mapping to 10 output classes.

\begin{table}[htbp]
  \centering
  \caption{CNNSmall architecture used for Fashion-MNIST classification.}
  \label{tab:cnnsmall-architecture}
  \begin{tabular}{llll}
    \hline
    \textbf{Layer} & \textbf{Operation} & \textbf{Parameters} & \textbf{Output Shape} \\
    \hline
    Input & Grayscale image & --- & $1 \times 28 \times 28$ \\
    Conv2D & 32 filters, $3 \times 3$ kernel, padding$=1$ & stride$=1$ & $32 \times 28 \times 28$ \\
    ReLU & Activation & --- & $32 \times 28 \times 28$ \\
    MaxPool2D & $2 \times 2$ & stride$=2$ & $32 \times 14 \times 14$ \\
    Conv2D & 64 filters, $3 \times 3$ kernel, padding$=1$ & stride$=1$ & $64 \times 14 \times 14$ \\
    ReLU & Activation & --- & $64 \times 14 \times 14$ \\
    MaxPool2D & $2 \times 2$ & stride$=2$ & $64 \times 7 \times 7$ \\
    Flatten & Reshape & $64 \times 7 \times 7 = 3136$ & $3136$ \\
    Linear & Fully connected & $3136 \rightarrow 10$ & 10 logits \\
    \hline
  \end{tabular}
\end{table}

The choice of a compact architecture was intentional. Limiting model capacity allows clearer observation of optimisation dynamics and reduces architectural complexity as a confounding factor in trajectory analysis. Since the objective of this work is diagnostic rather than performance-driven, architectural simplicity improves interpretability of gradient behaviour and loss evolution during PGD attacks. Trajectory-level assessments were therefore restricted to CNNSmall to ensure methodological consistency and to maintain a verified correspondence between trained weights and recorded adversarial trajectory metadata. This controlled setup prioritises diagnostic clarity over state-of-the-art accuracy.

\subsection{Training Procedure}
Models were trained using supervised learning with cross-entropy loss. Optimisation was performed using the Adam optimiser with a learning rate of $1 \times 10^{-3}$. Training was conducted for up to 75 epochs with a batch size of 128. Early stopping was implemented based on clean validation accuracy (patience = 5 for the clean-trained model and patience = 7 for adversarially trained models, using slightly higher patience for adversarial training to accommodate noisier validation dynamics and reduce premature stopping), and the best-performing checkpoint was restored to ensure stable convergence.

Adversarial training was performed using a 20-step Projected Gradient Descent attack (PGD-20). 
We trained \textbf{two separate adversarially trained models}: one with perturbation budget 
$\varepsilon = 0.1$ and one with $\varepsilon = 0.2$. 
For each model, adversarial examples were generated during training using the corresponding 
$\varepsilon$ value. The step size was set to $\alpha = 0.25\varepsilon$. 
The training objective combined clean and adversarial losses as:
\[
  \mathcal{L} = 0.5\,\mathcal{L}_{\mathrm{clean}} 
  + 0.5\,\mathcal{L}_{\mathrm{adv}}.
\]

Experiments were repeated across three random seeds $\{0,1,2\}$ to reduce sensitivity to initialization.

Classification accuracy on the unperturbed test set is reported as clean accuracy. Final clean and robust accuracies for each training regime are reported in Table~\ref{tab:robustness-results}. This metric serves both as a baseline robustness reference and as the conditioning criterion for trajectory-level analysis.

\subsection{Adversarial Attack: Projected Gradient Descent}

Adversarial examples were generated using Projected Gradient Descent (PGD), 
a standard iterative first-order white-box adversarial attack. 
PGD constructs adversarial inputs by performing gradient ascent on the 
cross-entropy loss with respect to the true label, followed by projection 
onto a constrained perturbation set defined by an $\ell_\infty$ norm bound.

In our experiments, an untargeted PGD attack was applied for $T = 20$ 
iterations under perturbation budgets $\varepsilon \in \{0.1, 0.2\}$. 
The step size was set proportionally to the perturbation radius, 
$\alpha = 0.25\varepsilon$, consistent with common robustness evaluation 
protocols. After each update, adversarial inputs were projected back 
onto the valid $\ell_\infty$ ball around the original image and clipped 
to the input range $[0,1]$.

The choice of $T = 20$ steps balances computational tractability with 
sufficient optimisation depth to approximate a strong first-order adversary. 
Empirically, loss trajectories saturated before the maximum iteration count 
across both perturbation budgets, indicating that 20 steps were sufficient 
to approximate convergence under the studied configuration.

\paragraph{Trajectory initialisation.}
During trajectory collection, we recorded a single optimisation path per sample.
For each clean-correct input $x^0$, PGD was initialised from a \emph{single random point} 
within the $\ell_\infty$ ball of radius $\varepsilon$ around $x^0$. 
Specifically, the initial adversarial iterate was defined as
$x^0 + \delta_0$, where $\delta_0 \sim \mathrm{Uniform}(-\varepsilon, \varepsilon)$,
followed by projection into the valid input domain.
After this initialisation, subsequent PGD updates were deterministic.
No multiple restarts were used during trajectory recording.

This design isolates one optimisation trajectory per sample, enabling 
step-by-step analysis of attack dynamics without conflating trajectory 
geometry with worst-case maximisation effects.

For robustness evaluation (Section~3.8), PGD-20 was performed with 
random initialisation and three restarts, with worst-case selection per example.

PGD under these settings is widely regarded as a strong first-order 
white-box benchmark and serves as the basis for both robustness 
evaluation and trajectory-level analysis in this work.

\paragraph{PGD update rule.}
Let $x^0$ denote the clean input and $y$ the true label.
At iteration $t$, PGD updates the adversarial iterate $x^t$ via
\[
  x^{t+1} =
  \Pi_{\mathcal{B}_\infty(x^0,\varepsilon)}
  \!\left(
    x^t + \alpha \, \mathrm{sign}
    \left(
      \nabla_x \mathcal{L}(f_\theta(x^t), y)
    \right)
  \right),
\]
where $\Pi_{\mathcal{B}_\infty(x^0,\varepsilon)}(\cdot)$ denotes projection 
onto the $\ell_\infty$ ball of radius $\varepsilon$ centred at $x^0$.

\subsection{PGD Trajectory Representation}

Our analysis explicitly records the entire trajectory of PGD for each input, unlike standard evaluation practice where only the final adversarial example is considered. We denote the trajectory of an adversarial example as a sequence
\[
  \{x^0, x^1, \dots, x^T\},
\]
where $x^0$ represents the original clean input and $x^t$ denotes the perturbed input after $t$ PGD updates. In particular, $x^T$ corresponds to the final adversarial example after $T$ iterations. Recording the full trajectory facilitates analysis of how loss values, gradients, and predictions evolve throughout the attack process rather than only at convergence.

For clarity, failure-step statistics reported later are indexed relative to adversarial updates only. Thus, a failure recorded at step 0 corresponds to misclassification at $x^1$ (i.e., immediately after the first PGD update), while the clean input $x^0$ is not counted as a failure step.

\subsection{Trajectory-Level Metrics}
To characterise adversarial dynamics, we calculate several metrics throughout the PGD trajectory.

\subsubsection{Loss Trajectory}
At each PGD step, we record the cross-entropy loss of the model on the perturbed input. Loss trajectories capture the progression of the attack objective and offer insights into the speed and stability of optimisation.

\subsubsection{Gradient Cosine Similarity}
To evaluate the stability of gradient directions during the attack steps, we calculate the cosine similarity between the gradients at successive PGD iterations. Gradient cosine similarity values lie in the range $[-1, 1]$, with values approaching 1 signifying stable update directions, values close to 0 indicating rapidly fluctuating gradients, and negative values reflecting oscillatory behaviour. This metric serves as an indicator of the local smoothness of the loss landscape experienced during the attack.

\subsubsection{Steps to Failure}
Steps-to-failure is defined as the initial PGD iteration where the model’s predicted label differs from the true label. This metric captures the point in the attack when classification failure occurs, providing a time-resolved view of the model’s robustness that extends beyond final accuracy.

\subsection{Data Filtering and the Clean-Correct Constraint}
Trajectory-level analysis is meaningful only when the model accurately categorises the clean input. Trajectory statistics may be skewed by trivially stable or degenerate behaviours from samples that are misclassified before any adversarial perturbation if they are not conditioned on clean correctness.

To ensure the integrity of the diagnostic signals, the analysis was strictly conditioned on clean-correct samples. For the retrained models, clean accuracy ranged between 84\% and 92\%, depending on the training regime. Trajectory collection was therefore performed on a subset of 3000 clean-correct test samples per model to ensure sufficient statistical power while maintaining computational tractability.

This filtering ensures that reported trajectory behaviours reflect adversarial vulnerability rather than pre-existing classification errors.

\subsection{Evaluation Protocol and Robust Accuracy}

Robustness is measured using \emph{robust accuracy}, defined as the proportion 
of test inputs that remain correctly classified under a fixed adversarial attack. 
Robust accuracy is reported alongside clean accuracy to contextualize 
robustness--accuracy trade-offs.

All robustness assessments were conducted using 20-step PGD attacks 
(PGD-20) under perturbation budgets $\varepsilon = 0.1$ and $\varepsilon = 0.2$. 
For robustness evaluation, PGD-20 included random initialisation within 
the $\ell_\infty$ ball and three independent restarts, with worst-case 
selection per example. This configuration approximates a strong 
first-order white-box adversary.

Under this evaluation protocol, the clean-trained model achieved 
$0/10000$ correct classifications for both $\varepsilon = 0.1$ and 
$\varepsilon = 0.2$, yielding robust accuracy of exactly $0.0000$ 
in both cases. Robust accuracy values are reported to four decimal 
places throughout.

Trajectory metrics are computed separately from robustness evaluation. 
While robustness evaluation uses multiple restarts to approximate 
worst-case performance, trajectory analysis records a single 
randomly initialised optimisation path per clean-correct sample 
(Section~3.4). All trajectory statistics are aggregated across 
samples using mean and standard deviation. Where applicable, 
shaded regions in plots represent inter-sample variability.

\subsection{Summary of Methodology}
In summary, the methodology extends standard PGD-based robustness evaluation by explicitly capturing and evaluating adversarial attack trajectories. This approach provides a more nuanced understanding of adversarial robustness beyond final accuracy alone through careful data conditioning and multiple trajectory-level indicators.

\section{Results and Discussion}
This section presents and interprets the empirical results of trajectory-level analyses of Projected Gradient Descent (PGD) attacks on three models trained on Fashion-MNIST: a clean-trained model, an adversarially trained model with $\varepsilon = 0.1$, and an adversarially trained model with $\varepsilon = 0.2$.

Robust evaluation was carried out using 20-step PGD with random initialization and multiple restarts. Trajectory-level metrics were computed on 3000 clean-correct samples per model.

Unless otherwise stated, all trajectory-level plots correspond to single-initialisation PGD-20 attacks under perturbation budget $\varepsilon = 0.1$. Although trajectories were recorded under both $\varepsilon \in \{0.1, 0.2\}$, we present $\varepsilon = 0.1$ results for the primary trajectory analysis for two reasons. First, $\varepsilon = 0.1$ represents a moderate perturbation regime in which all models exhibit non-trivial behaviour, enabling meaningful comparison of trajectory dynamics. Second, evaluating trajectories at the same perturbation budget used during adversarial training for one of the models provides a consistent baseline for analysing optimisation geometry across robustness regimes. 

Rather than relying merely on final adversarial accuracy, we investigate loss evolution, gradient alignment, and steps-to-failure to characterise adversarial optimisation dynamics and assess how well trajectory diagnostics distinguish robustness regimes.

The three models exhibit a clear robustness hierarchy under strong PGD evaluation, as summarized in Table~\ref{tab:robustness-results}. The clean-trained model collapses under attack, while the $\varepsilon = 0.1$ model exhibits moderate robustness. The $\varepsilon = 0.2$ model retains substantial robustness even at higher perturbation budgets. Trajectory-level behaviour is analysed in light of this hierarchy.

\begin{table}[H]
  \centering
  \small
  \setlength{\tabcolsep}{5pt}
  \renewcommand{\arraystretch}{1.2}
  \caption{Clean and robust accuracy under strong PGD evaluation for the clean-trained model and adversarially trained models at $\varepsilon = 0.1$ and $\varepsilon = 0.2$.}
  \label{tab:robustness-results}
  \begin{tabular}{lccc}
    \hline
    \textbf{Model} & \textbf{Clean Acc.} & \textbf{Robust Acc. ($\varepsilon = 0.1$)} & \textbf{Robust Acc. ($\varepsilon = 0.2$)} \\
    \hline
    Clean-trained & 0.9161 & 0.0000 & 0.0000 \\
    $\varepsilon = 0.1$ trained & 0.8602 & 0.7298 & 0.0835 \\
    $\varepsilon = 0.2$ trained & 0.8443 & 0.7526 & 0.6824 \\
    \hline
  \end{tabular}
\end{table}

As expected, robustness does not transfer across perturbation budgets.
The model adversarially trained with $\varepsilon = 0.1$ exhibits a substantial
performance drop when evaluated at $\varepsilon = 0.2$, decreasing from
$72.98\%$ robust accuracy at $\varepsilon = 0.1$ to $8.35\%$ at
$\varepsilon = 0.2$ (835/10000 correct).
This reflects the mismatch between the training and evaluation threat models
and is consistent with established observations that adversarial robustness
is highly budget-specific.

\subsection{Loss Evolution Along PGD Trajectories}

\begin{figure}[H]
  \centering
  \includegraphics[width=0.85\linewidth]{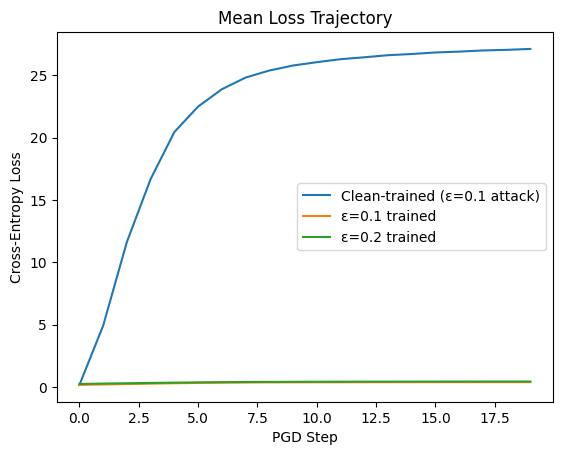}
  \caption{
    Mean cross-entropy loss across 20-step PGD trajectories for each model. The plot shows rapid loss escalation for the clean model, whereas both robust models exhibit nearly identical, suppressed loss growth.
  }
  \label{fig:loss-trajectory}
\end{figure}

Average cross-entropy loss across PGD iterations for each robustness regime is illustrated in Figure~\ref{fig:loss-trajectory}. All loss curves reported in this section are computed from recorded single-initialisation PGD-20 trajectories (without restarts) under perturbation budget $\varepsilon = 0.1$, consistent with the trajectory recording protocol described in the methods. For all models, the loss increases monotonically during early attack iterations until reaching saturation. This confirms that PGD performs gradient ascent on the loss under the provided $\ell_\infty$ restrictions.

The clean-trained model demonstrates rapid loss escalation, consistent with its near-zero robust accuracy under strong PGD evaluation. In comparison, both adversarially trained models demonstrate much slower and more controlled loss growth. Notably, the $\varepsilon = 0.1$ and $\varepsilon = 0.2$ adversarially trained
models exhibit quantitatively similar loss dynamics despite their substantially
different robust accuracies at $\varepsilon = 0.2$. 
The mean final-step loss values differ by only $0.052$ 
($0.388$ vs.\ $0.440$), corresponding to a relative difference of approximately 
$12.5\%$. Likewise, the area under the loss curve (AUC) across 20 steps differs 
by approximately $13.7\%$. 
In contrast, robust accuracy at $\varepsilon = 0.2$ differs by nearly 
$60$ percentage points ($8.35\%$ vs.\ $68.24\%$). 
This disparity indicates that mean loss magnitude and smoothness, while descriptive of optimisation behaviour, do not reliably distinguish robustness strength across regimes.

\subsection{Gradient Alignment and Local Optimisation Geometry}

\begin{figure}[H]
  \centering
  \includegraphics[width=0.85\linewidth]{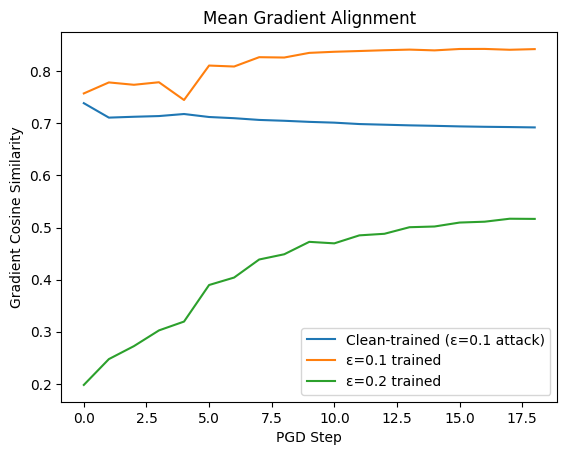}
  \caption{
    Mean cosine similarity between successive input gradients along the attack path. While the $\varepsilon=0.1$ model maintains high directional stability, the $\varepsilon=0.2$ model shows lower initial alignment that gradually stabilizes.
  }
  \label{fig:grad-cos-sim}
\end{figure}

Figure~\ref{fig:grad-cos-sim} reports cosine similarity between gradients at successive PGD steps, serving as a measure of directional stability along the optimisation trajectory. The clean-trained model demonstrates moderate initial alignment that diminishes across iterations, consistent with unstable ascent directions under attack.

The $\varepsilon = 0.1$ adversarially trained model exhibits consistently high gradient alignment throughout the trajectory, indicating coherent ascent directions in input space. In contrast, the $\varepsilon = 0.2$ model displays substantially lower initial alignment, followed by gradual stabilisation at later steps. This behaviour suggests a more heterogeneous local optimisation geometry at larger perturbation budgets before convergence to a stable ascent direction.

To quantify early-stage behaviour, we examine gradient cosine similarity at the first PGD step across samples. The $\varepsilon = 0.1$ model exhibits mean early-step alignment of $0.758 \pm 0.208$, whereas the $\varepsilon = 0.2$ model exhibits substantially lower and more variable alignment of $0.198 \pm 0.300$. Despite this reduced and more heterogeneous initial alignment, the $\varepsilon = 0.2$ model achieves markedly higher robust accuracy at $\varepsilon = 0.2$ ($68.24\%$ vs.\ $8.35\%$). Moreover, the overall alignment variance across steps is largest for the $\varepsilon = 0.2$ model.

These results indicate that strong adversarial robustness does not require consistently high or low-variance gradient alignment. Alignment smoothness characterises aspects of local optimisation geometry but does not uniquely determine functional resistance to adversarial perturbation. The progressive alignment increase observed for the $\varepsilon = 0.2$ model may reflect the attack converging toward a dominant ascent direction after early exploration of a flatter loss surface induced by training at a wider perturbation radius.

\subsection{Steps-to-Failure and Robustness Separation}

\begin{figure}[H]
  \centering
  \includegraphics[width=0.85\linewidth]{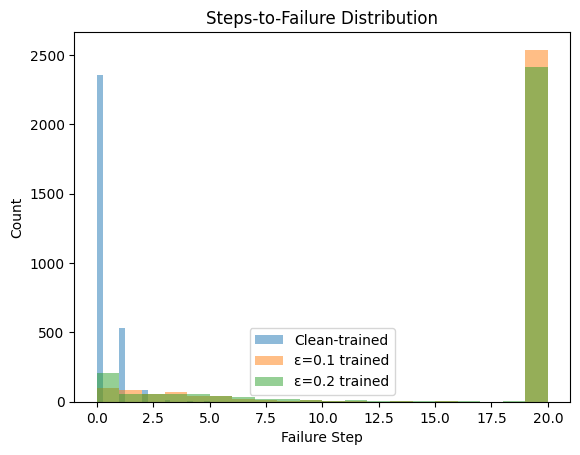}
  \caption{
    Distribution of first failure step during a 20-step PGD attack under perturbation budget $\varepsilon = 0.1$, the primary trajectory analysis budget used for cross-model comparison. This metric clearly separates the regimes, showing immediate failure for the clean model and a high survival rate for the $\varepsilon = 0.2$ model.
  }
  \label{fig:steps-to-failure}
\end{figure}

The distribution of steps-to-failure for clean-correct samples under a PGD attack is presented in Figure~\ref{fig:steps-to-failure}. This metric most clearly differentiates robustness regimes. The clean-trained model exhibits a sharp spike at step 0, corresponding to failure immediately after the first PGD update. This aligns with its near-zero robust accuracy.

The $\varepsilon = 0.1$ adversarially trained model reveals a scattered distribution: some samples fail early, while others resist perturbation for numerous steps. This generates a visible delay-to-failure structure consistent with intermediate robustness. The $\varepsilon = 0.2$ model exhibits a notable spike at the final PGD iteration, with a large number of samples surviving the complete attack.

Unlike mean loss trajectories, which appear deceptively comparable across adversarially trained models, the steps-to-failure distribution clearly differentiates robustness regimes. This shows that failure timing conveys functional robustness qualities more directly than smooth optimisation metrics.

\section{Conclusion}
This paper analyses when trajectory-level diagnostics of Projected Gradient Descent (PGD) attacks provide meaningful insights about adversarial robustness and when they may be insufficient if interpreted in isolation. We examined the entire trajectory of PGD using loss evolution, gradient alignment, and steps-to-failure as diagnostic signals rather than measuring robustness merely through final adversarial accuracy. Across clean-trained and adversarially trained models with varying robustness levels, our findings show that trajectory-level analysis offers substantial descriptive insight into adversarial optimisation dynamics; however, its interpretability depends critically on how these signals are contextualised.

\subsection{When Trajectory-Level Diagnostics Are Informative
}

Trajectory-level diagnostics are most informative when models operate within structured robustness regimes that exhibit meaningful variation in failure behaviour. In such contexts, steps-to-failure distributions provide a clear separation of robustness strength, describing how long inputs endure adversarial perturbation under strong PGD evaluation.

Gradient alignment and loss evolution further characterise the geometry of the optimisation trajectory, indicating how adversarial updates advance through input space. When interpreted collectively, these metrics provide a broader understanding of attack dynamics beyond final accuracy alone.

\subsection{When Trajectory-Level Diagnostics Become Insufficient}

However, our results reveal that certain trajectory metrics, particularly mean loss smoothness and gradient alignment, do not uniquely determine robustness strength. Notably, adversarially trained models with substantially different robust accuracies exhibited visually similar mean loss trajectories.

This demonstrates that smooth optimisation geometry does not necessarily correspond to similar adversarial resistance. While trajectory curves may appear stable and well-behaved, functional robustness differences are more clearly revealed through failure timing behaviour.

These findings indicate that trajectory-level signals characterise optimisation processes rather than directly measuring robustness. When examined in isolation, they may hide substantial variations among robustness regimes.

\subsection{Implications for Adversarial Robustness Evaluation}

The consequences of this work extend to the broader practice of robustness evaluation. As trajectory-level diagnostics become increasingly common, there is a risk of confusing smooth or steady optimisation behaviour with resilience strength.

Our findings show that trajectory metrics should not replace standard robustness assessments. Instead, they should be viewed as supplementary descriptions of adversarial dynamics, interpreted alongside robust accuracy and failure distributions under strong attack settings.

Multi-metric evaluation and clear reporting of attack configurations are vital for preventing over-interpretation of optimisation geometry.

\subsection{Limitations and Future Directions}

This study relies on a single dataset (Fashion-MNIST) and a single compact architecture (CNNSmall). Although the retrained models demonstrate realistic robustness trade-offs, the outcomes should be understood within this experimental environment rather than as universal claims.

While PGD-20 with random initialisation and multiple restarts provides a strong first-order evaluation baseline, future work may incorporate parameter-free ensembles such as AutoAttack to further strengthen robustness validation under complementary attack strategies. Further exploration into formal criteria for interpreting trajectory-level measurements may also help establish principled safeguards against metric over-interpretation.

\subsection{Closing Remarks}

Trajectory-level analysis provides substantial insight into adversarial optimisation geometry; however, it does not, by itself, fully reflect adversarial robustness. Functional resistance to attack is more directly captured by failure timing behaviour than by smoothness of optimisation trajectories.

By clarifying both the strengths and the limits of trajectory-level diagnostics under well-trained robustness regimes, this work offers a more precise perspective on their role in adversarial robustness evaluation.

\bibliographystyle{abbrv}
\bibliography{references}

@article{szegedy2013intriguing,
  title={Intriguing properties of neural networks},
  author={Szegedy, Christian and Zaremba, Wojciech and Sutskever, Ilya and Bruna, Joan and Erhan, Dumitru and Goodfellow, Ian and Fergus, Rob},
  journal={arXiv preprint arXiv:1312.6199},
  year={2013}
}

@article{goodfellow2014explaining,
  title={Explaining and harnessing adversarial examples},
  author={Goodfellow, Ian J and Shlens, Jonathon and Szegedy, Christian},
  journal={arXiv preprint arXiv:1412.6572},
  year={2014}
}

@article{madry2017towards,
  title={Towards deep learning models resistant to adversarial attacks},
  author={Madry, Aleksander and Makelov, Aleksandar and Schmidt, Ludwig and Tsipras, Dimitris and Vladu, Adrian},
  journal={arXiv preprint arXiv:1706.06083},
  year={2017}
}

@article{ilyas2019adversarial,
  title={Adversarial examples are not bugs, they are features},
  author={Ilyas, Andrew and Santurkar, Shibani and Tsipras, Dimitris and Engstrom, Logan and Tran, Brandon and Madry, Aleksander},
  journal={Advances in neural information processing systems},
  volume={32},
  year={2019}
}

@inproceedings{athalye2018obfuscated,
  title={Obfuscated gradients give a false sense of security: Circumventing defenses to adversarial examples},
  author={Athalye, Anish and Carlini, Nicholas and Wagner, David},
  booktitle={Proceedings of the 35th International Conference on Machine Learning},
  pages={274--283},
  year={2018},
  organization={PMLR},
  volume={80}
}

@inproceedings{croce2020reliable,
  title={Reliable evaluation of adversarial robustness with an ensemble of diverse parameter-free attacks},
  author={Croce, Francesco and Hein, Matthias},
  booktitle={Proceedings of the 37th International Conference on Machine Learning},
  series={Proceedings of Machine Learning Research},
  pages={2206--2216},
  year={2020},
  organization={PMLR},
  volume={119}
}

@article{liu2020loss,
  title={On the loss landscape of adversarial training: Identifying challenges and how to overcome them},
  author={Liu, Chen and Salzmann, Mathieu and Lin, Tao and Tomioka, Ryota and S{\"u}sstrunk, Sabine},
  journal={Advances in Neural Information Processing Systems},
  volume={33},
  pages={21476--21487},
  year={2020}
}

@article{Xiao2017FashionMNISTAN,
  title={Fashion-MNIST: a Novel Image Dataset for Benchmarking Machine Learning Algorithms},
  author={Han Xiao and Kashif Rasul and Roland Vollgraf},
  journal={arXiv: Computer Vision and Pattern Recognition},
  year={2017},
  url={https://github.com/zalandoresearch/fashion-mnist},
  doi={10.48550/arXiv.1708.07747}
}

@inproceedings{dong2018momentum,
  title={Boosting Adversarial Attacks with Momentum},
  author={Dong, Yinpeng and Liao, Fangzhou and Pang, Tianyu and Su, Hang and Zhu, Jun and Hu, Xiaolin and Li, Jianguo},
  booktitle={Proceedings of the IEEE Conference on Computer Vision and Pattern Recognition (CVPR)},
  year={2018}
}

\end{document}